\documentclass{article} 
\usepackage{nips12submit_e,times}

\usepackage{times}
\usepackage{graphicx}
\usepackage{amsmath}
\usepackage{amssymb}
\usepackage{subfigure}
\usepackage{multirow}
\usepackage{multicol}
\title{Indoor Semantic Segmentation\\ using depth information}

\author{
Camille Couprie$^1$\footnotemark[1]
\and
{\bf Cl\'ement Farabet}$^{2,3}$ 
\and
{\bf Laurent Najman}$^3$
\and
{\bf Yann LeCun}$^2$
\and
\\
$^1$ IFP Energies Nouvelles\\
Technology, Computer Science and Applied Mathematics Division\\
 Rueil Malmaison, France
\and
\\
$^2$ Courant Institute of Mathematical Sciences\\
New York University\\
New York, NY 10003, USA
\and
\\
$^3$  Universit\'e Paris-Est\\
 Laboratoire d'Informatique Gaspard-Monge\\ 
\'Equipe A3SI - ESIEE Paris, France
}

\usepackage{url}
\usepackage[backref,colorlinks=true]{hyperref}

\newcommand{\omitme}[1]{}

\nipsfinalcopy 

\begin{document}

\maketitle

\begin{abstract}
This work addresses multi-class segmentation of indoor scenes with
RGB-D inputs. While this area of research has gained much attention
recently, most works still rely on hand-crafted features. In contrast,
we apply a multiscale convolutional network to learn features directly
from the images and the depth information. We obtain state-of-the-art on the NYU-v2 depth dataset
with an accuracy of 64.5\%. We illustrate the labeling of indoor
scenes in videos sequences that could be processed in real-time using
 appropriate hardware such as an FPGA.

  \omitme{The task of associating a semantic class to the objects present in
  an image involves the joint segmentation and recognition of the
  objects and is thus a is challenging problem. In indoor
  environments, fortunately, depth information may be collected. This
  extra information started being exploited in computer visions
  applications. For instance, the NYU depth dataset collects diverse
  indoor scenes images with their depth information, and is the first
  dataset that also contains densely semantically labeled images.  In
  this work, we apply a multiscale convolutional network to learn
  features directly from the images and the depth information. The
  exploitation of a new image modality such as depth is
  straightforward when employing a feature learning approach. The
  results obtained on the NYU depth v2 dataset constitute the
  state-of-the-art with an accuracy of 64.5\%. We illustrate the
  labeling of indoor scenes in videos sequences that could be
  processed in real-time using appropriate hardware.}
\end{abstract}

\footnotetext[1]{$^*$ Performed the work at New York University.}

\section{Introduction}

The recent release of the Kinect allowed many progress in indoor
computer vision.  Most approaches have focused on object recognition
\cite{berkeley, Janoch11ICCVkinectworkshop} or point cloud semantic labeling \cite{cornell}, finding their
applications in robotics or games \cite{cruz2012kinectTutoriel}.  The
pioneering work of Silberman {\it et al.}
\cite{Silberman2011ICCVworkshop} was the first to deal with the task of semantic full image labeling using depth
information. The NYU depth v1 dataset \cite{Silberman2011ICCVworkshop}
guathers 2347 triplets of images, depth maps, and ground truth
labeled images covering twelve object categories. Most datasets
employed for semantic image segmentation
\cite{gould09_iccv,liu2010} present the objects centered into
the images, under nice lightening conditions. The NYU depth dataset
aims to develop joint segmentation and classification solutions to an
environment that we are likely to encounter in the everyday life. This
indoor dataset contains scenes of offices, stores, rooms of houses
containing many occluded objects unevenly lightened.
The first results \cite{Silberman2011ICCVworkshop} on this dataset were obtained using the extraction of sift features on the depth maps in addition to the RGB images. The depth
is then used in the gradient information to refine the predictions using graph cuts. Alternative CRF-like approaches have also been
explored to improve the computation time performances \cite{Coup1208:Multi}.
The results on NYU dataset v1 have been improved by \cite{Ren2012NYUdepthCVPR} using elaborate kernel descriptors and a post-processing step that employs gPb superpixels MRFs, involving large computation times.

A second version of the NYU depth dataset was released more recently \cite{Silberman:ECCV12}, 
and improves the labels categorization into 894 different object classes.
Furthermore, the size of the dataset did also increase, it now contains hundreds of video sequences
(407024 frames) acquired with depth maps.

Feature learning, or deep learning approaches are particularly 
adapted to the addition of new image
modalities such as depth information. Its recent success for dealing
with various types of data is manifest in speech recognition \cite{speech},
molecular activity prediction, object recognition \cite{imagenet} and many more applications. In computer
vision, the approach of Farabet {\it et al.} \cite{FarabetICML2012,FarabetPAMI} has
been specifically designed for full scene labeling and has proven its
efficiency for outdoor scenes. The key idea is to learn hierarchical
features by the mean of a multiscale convolutional network. Training networks using multiscales representation appeared also the same year in \cite{Ciresan:2012f,Schulz2012learning}.
 
When the depth information was not yet available, there have been
attempts to use stereo image pairs to improve the feature learning of
convolutional networks \cite{Lecun2004norb}. Now that depth maps are
easy to acquire, deep learning approachs started to be considered for improving object recognition \cite{SocherEtAl2012:CRNN}. In this work, we suggest to adapt Farabet {\it et al.}'s network to
learn more effective features for indoor scene labeling.  Our work is,
to the best of our knowledge, the first exploitation of depth
information in a feature learning approach for full scene labeling.

\section{Full scene labeling}
\label{sec:full}

\subsection{Multi-scale feature extraction}

Good internal representations are hierarchical. In vision,
pixels are assembled into edglets, edglets into motifs,
motifs into parts, parts into objects, and objects into
scenes. This suggests that recognition architectures for
vision (and for other modalities such as audio and
natural language) should have multiple trainable stages
stacked on top of each other, one for each level in the
feature hierarchy. Convolutional Networks \cite{Lecun1998gradient} (ConvNets)
provide a simple framework to learn such hierarchies of
features.

\omitme{In this work, we use the state-of-the-art feature learning system
for scene labeling, naming the multiscale convolutional network of
Farabet {\it et al.} \cite{FarabetPAMI}.}

Convolutional Networks are trainable
architectures composed of multiple stages. The input and output of
each stage are sets of arrays called feature maps.  In our case, the
input is a color (RGB) image plus a depth (D) image and each feature map
is a 2D array containing a color or depth channel of the input RGBD
image. At the output, each feature map represents a particular feature
extracted at all locations on the input. Each stage is composed of
three layers: a filter bank layer, a non-linearity layer, and a
feature pooling layer. A typical ConvNet is composed of one, two or
three such 3-layer stages, followed by a classification
module. Because they are trainable, arbitrary input modalities can be
modeled, such as the depth modality that is added to the input channel
in this work.

\begin{figure}[htb]
\includegraphics[scale=0.081]{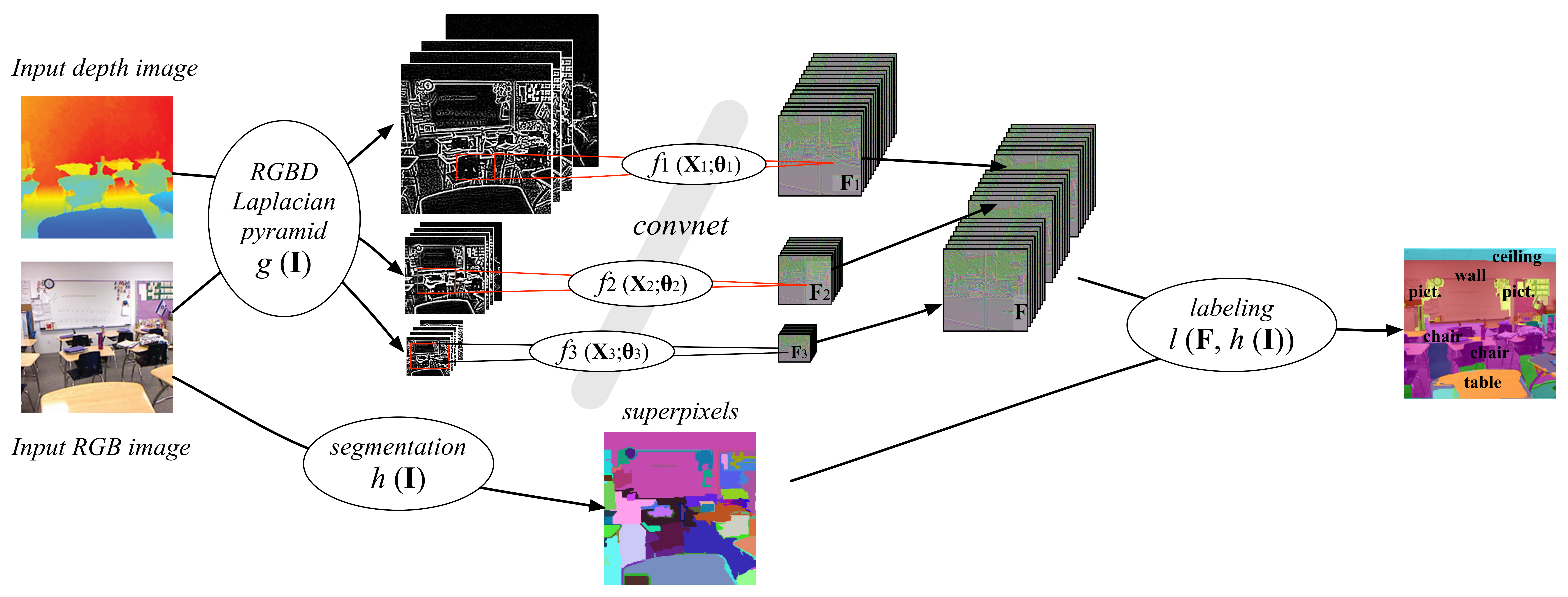}
\caption{Scene parsing (frame by frame) using a multiscale network and
  superpixels. The RGB channels of the image and the depth image are
  transformed through a Laplacian pyramid. Each scale is fed to a
  3-stage convolutional network, which produces a set of feature
  maps. The feature maps of all scales are concatenated, the
  coarser-scale maps being upsampled to match the size of the
  finest-scale map. Each feature vector thus represents a large
  contextual window around each pixel. In parallel, a single
  segmentation of the image into superpixels is computed
  to exploit the natural contours of the image. The final labeling is
  obtained by the aggregation of the classifier predictions into the
  superpixels.}
\label{model}
\end{figure}

A great gain has been achieved with the introduction of the {\it multiscale}
convolutional network described in \cite{FarabetPAMI}. The
multi-scale, dense feature extractor produces a series of feature
vectors for regions of multiple sizes centered around every pixel in
the image, covering a large context. The multi-scale convolutional net
contains multiple copies of a single network \omitme{(all sharing the same
weights)} that are applied to different scales of a Laplacian pyramid
version of the RGBD input image.

The RGBD image is first pre-processed, so that local neighborhoods
have zero mean and unit standard deviation. The depth image, given in
meters, is treated as an additional channel similarly to any color
channel. The overview scheme of our model appears in Figure \ref{model}.

Beside the input image which is now including a depth channel, the
parameters of the multi-scale network (number of scales, sizes of
feature maps, pooling type, etc.) are identical to
\cite{FarabetPAMI}. The feature maps sizes are 16,64,256, multiplied
by the three scales. The size of convolutions kernels are set to 7 by 7 at
each layer, and sizes of subsampling kernels (max pooling) are 2 by
2. In our tests we rescaled the images to the size $240 \times 320$.

As in \cite{FarabetPAMI}, the feature
extractor followed by a classifier was trained to minimize the
negative log-likelihood loss function. 
The classifier that follows feature extraction is a 2-layer
multi-perceptron, with a hidden layer of size 1024. 
We use superpixels \cite{Felzenszwalb04efficientgraph-based} to smooth the convnet predictions as a post-processing step, by agregating the classifiers predictions in each superpixel.

\omitme{ The output of the multiscale convnet was $256\times3 =
768$ feature maps of size $60\times80$. 

 Instead of only using the RGB
channel of the images, we add a fourth channel corresponding to the
depth information, given in meters.}

\subsection{Movie processing}

While the training is performed on single images, we are able to
perform scene labeling of video sequences. In order to improve the
performances of our frame-by-frame predictions, a temporal smoothing
may be applied. In this work, instead of using the frame by frame
superpixels as in the previous section, we employ the temporal
consistent superpixels of \cite{couprieCVPR2013}. This approach works
in quasi-linear time and reduces the flickering of objects that may
appear in the video sequences.

\section{Results}

We used for our experiments the NYU depth dataset -- version 2 -- of Silberman and
Fergus~\cite{Silberman:ECCV12}, composed of 407024 couples of
RGB images and depth images. Among these images, 1449 frames have been labeled. The object labels cover 894 categories. The dataset is provided with the original raw depth data that contain missing values, with code using \cite{Dani04colorizationusing} to inpaint the depth images.  

\subsection{Validation on images}

The training has been performed using the 894 categories directly as output classes. The frequencies of object appearences have not been changed in the training process. 
However, we established 14 clusters of classes categories to evaluate our
results more easily. The distributions of number of pixels per class
categories are given in Table \ref{classes_cat}.
We used the train/test splits as provided by the NYU depth v2 dataset, that is to say 795 training images and  654 test images.
Please note that no jitter (rotation, translations or any other transformation) was added to the dataset to gain extra performances. 
However, this strategy could be employed in future work.
The code consists of Lua scripts using the Torch machine learning software \cite{torch} available online at \begin{url}http://www.torch.ch/ \end{url}.

To evaluate the influence of the addition of depth information, we
trained a multiscale convnet only on the RGB channels, and another
network using the additional depth information. Both networks were trained until
the achievement of their best performances, that is to say for 105
epochs and 98 epochs respectively, taking less than 2 days on a regular server.

We report in Table \ref{classes_cat} two different performance measures: \begin{itemize}
 \item the ``classwise accuracy'', counting the number of correctly
   classified pixels divided by the number of false positive, averaged
   for each class. This number corresponds to the mean of the
   confusion matrix diagonal.
\item the ``pixelwise accuracy'', counting the number of correctly
  classified pixels divided by the total number of pixels of the test
  data.
\end{itemize}

\begin{figure*}[htb]
\begin{center}

\includegraphics[width=0.18\textwidth]{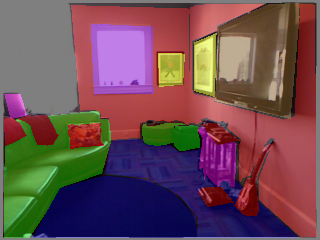}~
\includegraphics[width=0.18\textwidth]{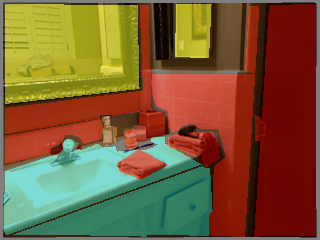}~
\includegraphics[width=0.18\textwidth]{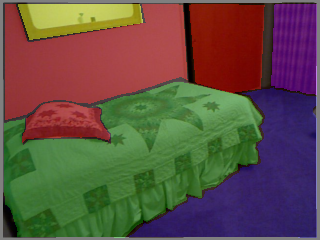}~
\includegraphics[width=0.18\textwidth]{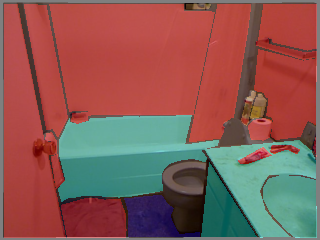}~
\includegraphics[width=0.18\textwidth]{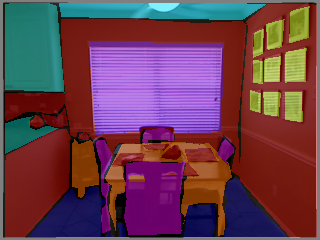}\\
 Ground truths \\
\includegraphics[width=0.18\textwidth]{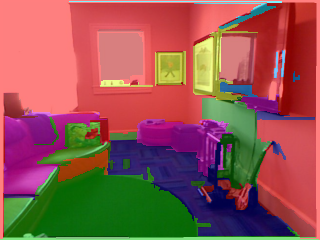}~
\includegraphics[width=0.18\textwidth]{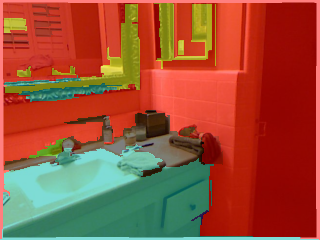}~
\includegraphics[width=0.18\textwidth]{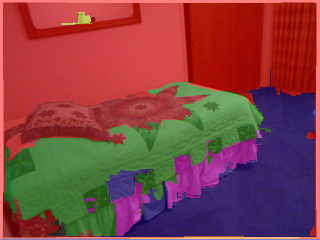}~
\includegraphics[width=0.18\textwidth]{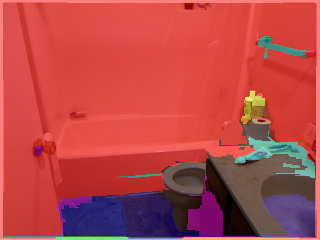}~
\includegraphics[width=0.18\textwidth]{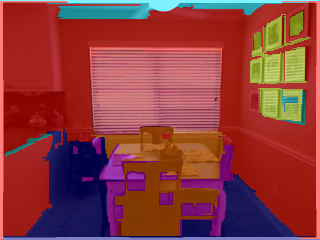}\\
Results using the Multiscale Convnet   \\
\includegraphics[width=0.18\textwidth]{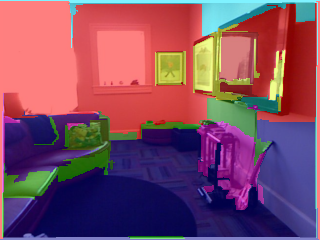}~
\includegraphics[width=0.18\textwidth]{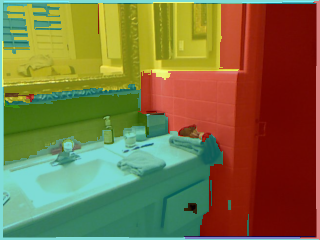}~
\includegraphics[width=0.18\textwidth]{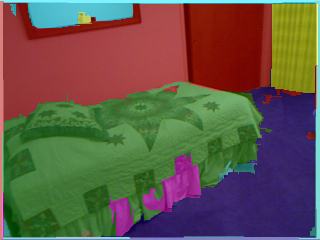}~
\includegraphics[width=0.18\textwidth]{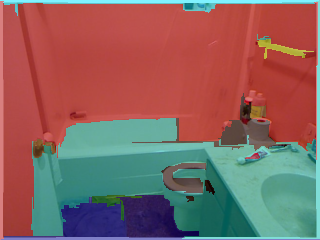}~
\includegraphics[width=0.18\textwidth]{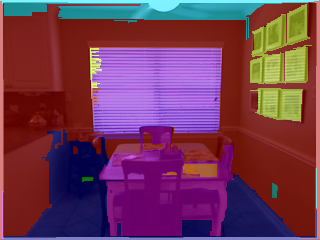}\\
Results using the Multiscale Convnet with depth information \\
\bigskip 
{\small
\begin{tabular}{cccccccc}
\begin{tabular}{l} 
\includegraphics[width=0.025\textwidth]{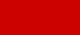} wall\\
\includegraphics[width=0.025\textwidth]{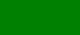} bed
\end{tabular}&
\begin{tabular}{l}
\includegraphics[width=0.025\textwidth]{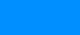} books \\
\includegraphics[width=0.025\textwidth]{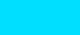} ceiling
\end{tabular}&
\begin{tabular}{l}
\includegraphics[width=0.025\textwidth]{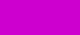} chair \\
\includegraphics[width=0.025\textwidth]{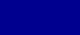} floor
\end{tabular}&
\begin{tabular}{l}
\includegraphics[width=0.025\textwidth]{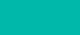} furniture \\
\includegraphics[width=0.025\textwidth]{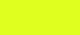} pict./deco 
\end{tabular}&
\begin{tabular}{l}
\includegraphics[width=0.025\textwidth]{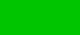} sofa\\
\includegraphics[width=0.025\textwidth]{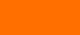} table
\end{tabular}&
\begin{tabular}{l}
\includegraphics[width=0.025\textwidth]{figure2k.png} object  \\
\includegraphics[width=0.025\textwidth]{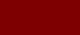} window\\
\end{tabular}&
\begin{tabular}{l}
\includegraphics[width=0.025\textwidth]{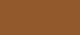} TV  \\
\includegraphics[width=0.025\textwidth]{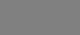} uknw\\
\end{tabular}&
\end{tabular}}\\
\bigskip
\includegraphics[width=0.18\textwidth]{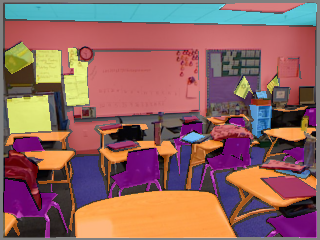}~
\includegraphics[width=0.18\textwidth]{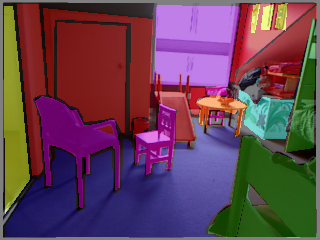}~
\includegraphics[width=0.18\textwidth]{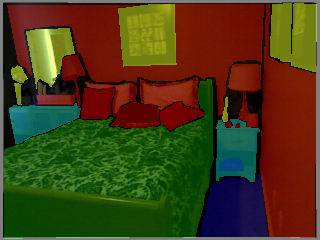}~
\includegraphics[width=0.18\textwidth]{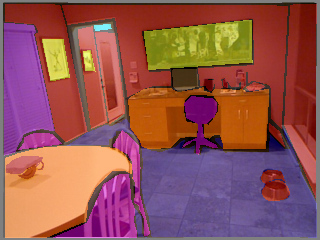}~
\includegraphics[width=0.18\textwidth]{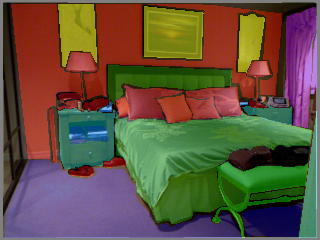}\\
Ground truths \\
\includegraphics[width=0.18\textwidth]{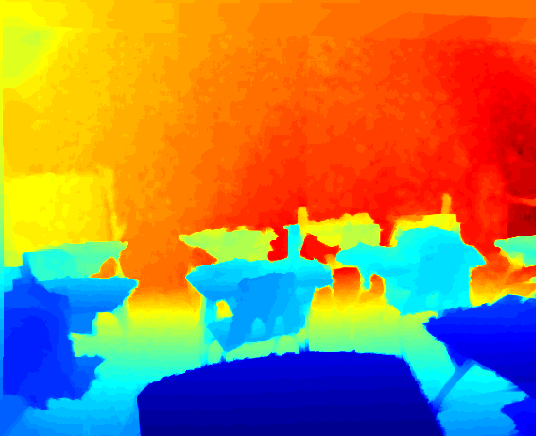}~
\includegraphics[width=0.18\textwidth]{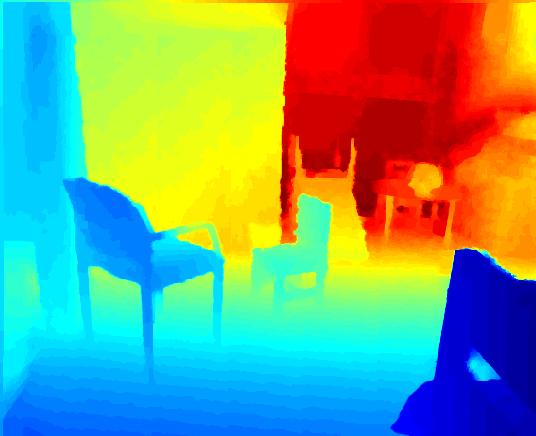}~
\includegraphics[width=0.18\textwidth]{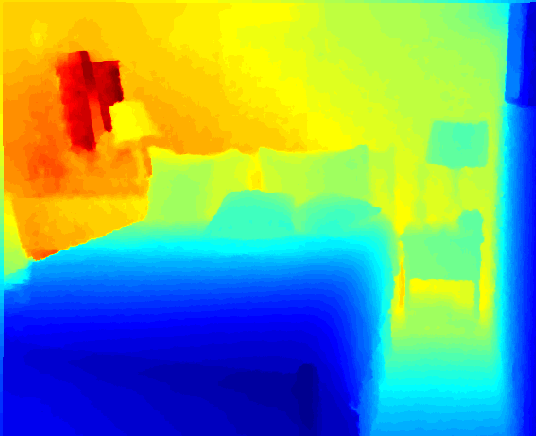}~
\includegraphics[width=0.18\textwidth]{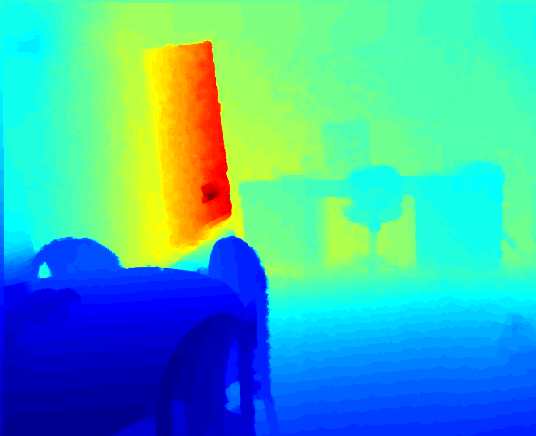}~
\includegraphics[width=0.18\textwidth]{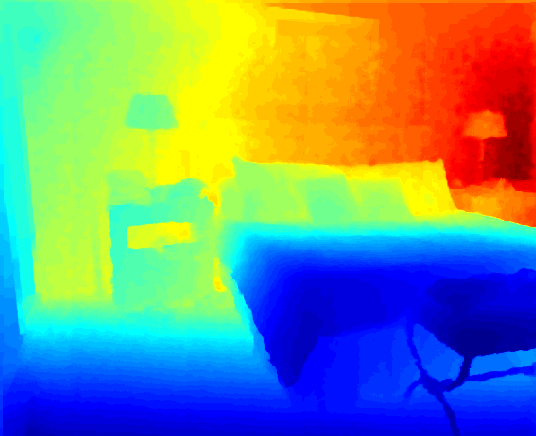}\\
Depth maps \\
\includegraphics[width=0.18\textwidth]{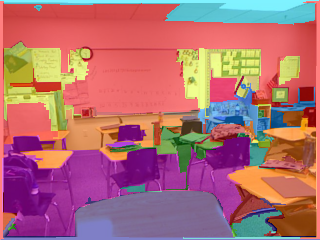}~
\includegraphics[width=0.18\textwidth]{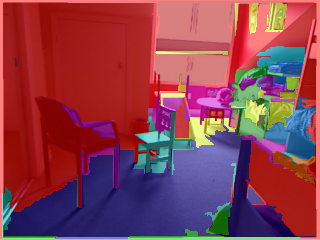}~
\includegraphics[width=0.18\textwidth]{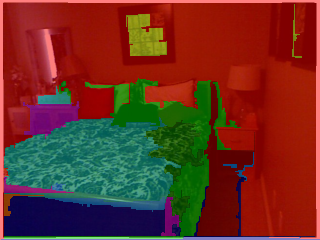}~
\includegraphics[width=0.18\textwidth]{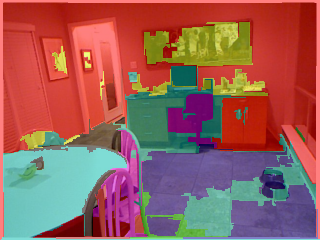}~
\includegraphics[width=0.18\textwidth]{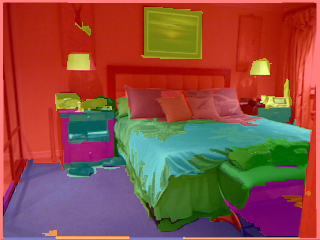}\\
Results using the Multiscale Convnet   \\
\includegraphics[width=0.18\textwidth]{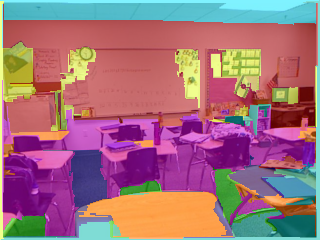}~
\includegraphics[width=0.18\textwidth]{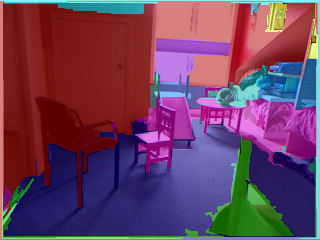}~
\includegraphics[width=0.18\textwidth]{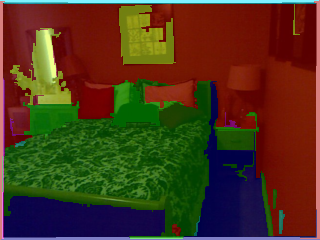}~
\includegraphics[width=0.18\textwidth]{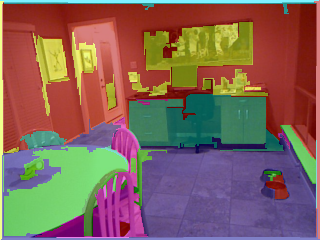}~
\includegraphics[width=0.18\textwidth]{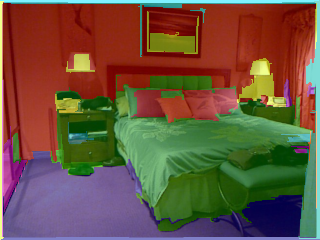}\\
Results using the Multiscale Convnet with depth information  \\
\end{center}
\caption{Some scene labelings using our Multiscale Convolutional Network trained on RGB and RGBD images. We observe in Table~\ref{classes_cat} that adding depth information helps to recognize objects that have low intra-class variance of depth appearance.}
\label{fig:results}
\end{figure*}

We observe that considerable gains (15\% or more) are achieved for the
classes 'floor', 'ceiling', and 'furniture'.  This result makes a lot
of sense since these classes are characterized by a somehow constant
appearance of their depth map. Objects such as TV, table, books can either be located in
the foreground as well as in the background of images. On the
contrary, the floor and ceiling will almost always lead to a depth
gradient always oriented in the same direction: Since the dataset has been collected by a person holding a
kinect device at a his chest, floors and ceiling are located at a
distance that does not vary to much through the dataset. Figure \ref{fig:results} provides examples of depth maps that illustrate these observations.
Overall, improvements induced by the depth information exploitation are present. In the next section, these improvements are more apparent. 

\begin{table}
\begin{center}
\begin{tabular}{c|c||c|c|}
\cline{2-4}
     &\multicolumn{1}{|c|}{Class} &\multicolumn{1}{|c|}{Multiscale} & \multicolumn{1}{|c|}{MultiScl. Cnet}  \\
     &\multicolumn{1}{|c|}{Occurrences} &\multicolumn{1}{|c|}{Convnet Acc. \cite{FarabetPAMI}} & \multicolumn{1}{|c|}{+depth Acc.} \\
\cline{2-4}
\hline
 \multicolumn{1}{|l|}{bed} &  4.4\%&   30.3    &{\bf 38.1} \\
 \multicolumn{1}{|l|}{objects} &7.1 \%&  {\bf 10.9}     &8.7\\
 \multicolumn{1}{|l|}{chair} &3.4\% &     {\bf 44.4}     &34.1\\
 \multicolumn{1}{|l|}{furnit.} &12.3\% &  28.5       & {\bf 42.4}\\
 \multicolumn{1}{|l|}{ceiling} & 1.4\% &   33.2       &{\bf 62.6}\\
 \multicolumn{1}{|l|}{floor} &9.9\% &     68.0       &{\bf 87.3}\\
\multicolumn{1}{|l|}{deco.} &3.4\% &     38.5       &{\bf 40.4}\\
\multicolumn{1}{|l|}{ sofa} & 3.2\% &     {\bf 25.8}      & 24.6\\
\multicolumn{1}{|l|}{ table} & 3.7\% &    {\bf 18.0}       & 10.2\\
\multicolumn{1}{|l|}{ wall} &24.5\% &     {\bf 89.4}       & 86.1 \\
\multicolumn{1}{|l|}{ window} & 5.1\% &   {\bf  37.8}     & 15.9 \\
\multicolumn{1}{|l|}{ books} &2.9\% &    {\bf 31.7}       & 13.7 \\
\multicolumn{1}{|l|}{ TV} &1.0\% &      {\bf  18.8}       & 6.0 \\
\multicolumn{1}{|l|}{ unkn.} & 17.8\% & -          & -\\
\hline
\hline
 \multicolumn{1}{|l|}{}  &  &   & \\
\multicolumn{1}{|l|}{ Avg. Class Acc.} & - & 35.8 & {\bf 36.2}\\
\hline
\hline
\multicolumn{1}{|l|}{ Pixel Accuracy (mean)} & - & 51.0   & {\bf 52.4} \\
\hline
\multicolumn{1}{|l|}{ Pixel Accuracy (median)} & - & 51.7   & {\bf 52.9} \\
\hline
\multicolumn{1}{|l|}{ Pixel Accuracy (std. dev.)} & - & 15.2   &  15.2 \\
\hline
\end{tabular}
\end{center}
\caption{Class occurrences in the test set -- Performances per class and per pixel.}
\label{classes_cat}
\end{table}


\subsection{Comparison with Silberman et al.}

In order to compare our results to the state-of-the-art on the NYU
depth v2 dataset, we adopted a different selection of outputs instead
of the 14 classes employed in the previous section.  The work of
Silberman {\it et al.} \cite{Silberman:ECCV12} defines the four
semantic classes Ground, Furniture, Props and Structure. This class selection is adopted in \cite{Silberman:ECCV12} to use semantic labelings of scenes to infer support relations between objects.  We recall
that the recognition of the semantic categories is performed in
\cite{Silberman:ECCV12} by the definition of diverse features
including SIFT features, histograms of surface normals, 2D and 3D
bounding box dimensions, color histograms, and relative depth.
 
\begin{table}[ht]
\begin{center}
\begin{tabular}{|c|c|c|c|c||c||c|}
\hline
 & Ground & Furniture & Props & Structure & Class Acc. & Pixel Acc. \\
\hline
Silberman {\it et al.}\cite{Silberman:ECCV12} & 68 & {\bf 70} & {\bf 42} & 59 & 59.6 & 58.6 \\
\hline
Multiscale convnet \cite{FarabetPAMI} &                         68.1 & 51.1 & 29.9 & {\bf 87.8} & 59.2 & 63.0  \\
\hline
Multiscale+depth convnet &                {\bf 87.3} & 45.3 & 35.5 & 86.1 & {\bf 63.5} & {\bf 64.5} \\
\hline
\end{tabular}
\end{center}
\caption{Accuracy of the multiscale convnet compared with the state-of-the-art approach of \cite{Silberman:ECCV12}.}
\label{classes_perfs}
\end{table}

As reported in Table \ref{classes_perfs}, the results achieved using
the Multiscale convnet are improving the structure class predictions,
resulting in a 4\% gain in pixelwise accuracy over Silberman {\it et
  al.} approach.  Adding the depth information results in a
considerable improvement of the ground prediction, and performs also
better over the other classes, achieving a 4\% gain in classwise
accuracy over previous works and improves by almost 6\% the pixelwise
accuracy compared to Silberman {\it et al.}'s results.

We note that the class 'furniture' in the 4-classes evaluation is
different than the 'furniture' class of the 14-classes evaluation. The
furniture-4 class encompasses chairs and beds but not desks, and
cabinets for example, explaining a drop of performances here using the
depth information.

\subsection{Test on videos}

The NYU v2 depth dataset contains several hundreds of video sequences
encompassing 26 different classes of indoor scenes, going from
bedrooms to basements, and dining rooms to book stores. Unfortunately,
no ground truth is yet available to evaluate our performances on this
video. Therefore, we only present here some illustrations of the
capacity of our model to label these scenes.

\begin{figure*}[htb]
\begin{center}
\includegraphics[width=0.16\textwidth]{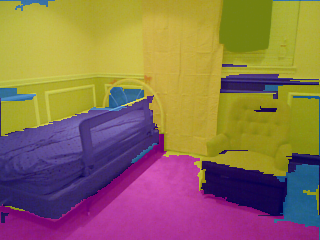}~
\includegraphics[width=0.16\textwidth]{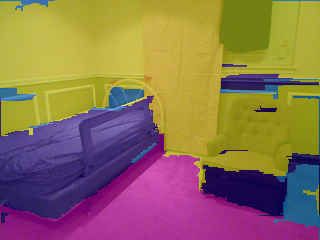}~
\includegraphics[width=0.16\textwidth]{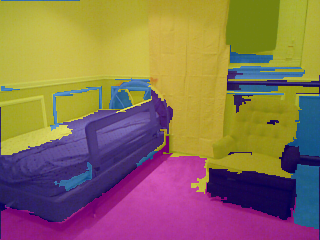}~
\includegraphics[width=0.16\textwidth]{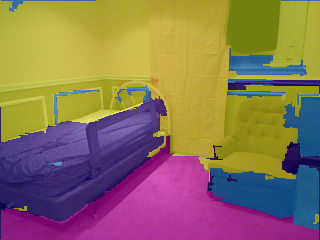}~
\includegraphics[width=0.16\textwidth]{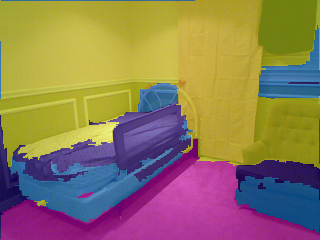}~
\includegraphics[width=0.16\textwidth]{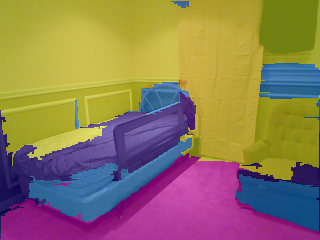}\\
(a) Output of the Multiscale convnet trained using depth information - frame by frame  \\
\includegraphics[width=0.16\textwidth]{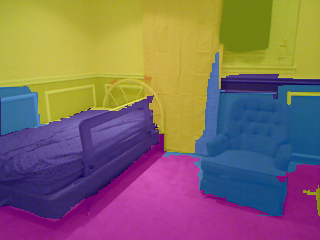}~
\includegraphics[width=0.16\textwidth]{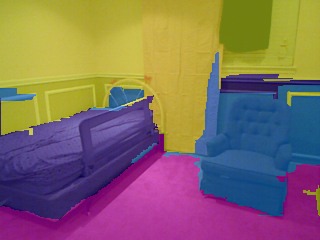}~
\includegraphics[width=0.16\textwidth]{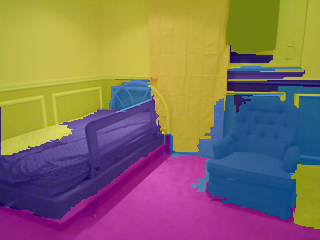}~
\includegraphics[width=0.16\textwidth]{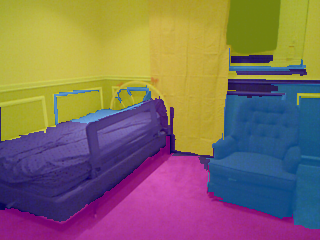}~
\includegraphics[width=0.16\textwidth]{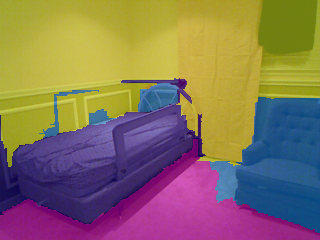}~
\includegraphics[width=0.16\textwidth]{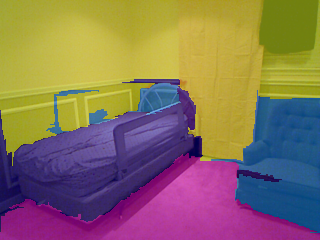}\\
(b) Results smoothed temporally using \cite{couprieCVPR2013}\\
\bigskip 
\begin{tabular}{cccc}
\includegraphics[width=0.025\textwidth]{figure2m.png} Props &
\includegraphics[width=0.025\textwidth]{figure2o.png} Floor &
\includegraphics[width=0.025\textwidth]{figure2p.png} Structure &
\includegraphics[width=0.025\textwidth]{figure2r.png} Wall 
\end{tabular}
\bigskip
\includegraphics[width=0.16\textwidth]{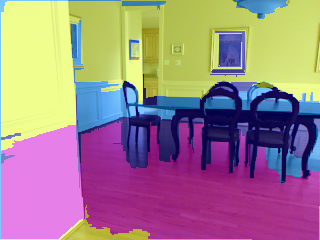}~
\includegraphics[width=0.16\textwidth]{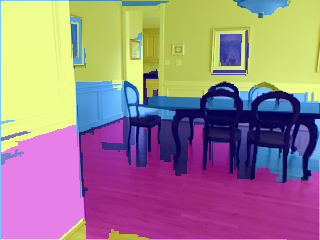}~
\includegraphics[width=0.16\textwidth]{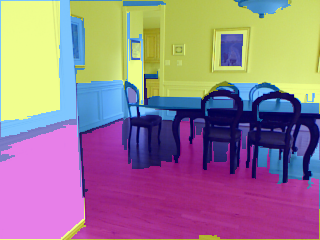}~
\includegraphics[width=0.16\textwidth]{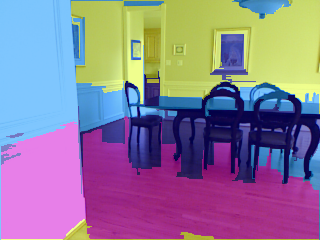}~
\includegraphics[width=0.16\textwidth]{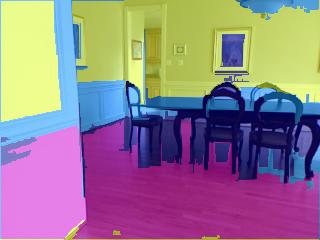}~
\includegraphics[width=0.16\textwidth]{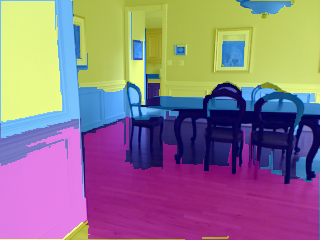}\\
(c) Output of the Multiscale convnet trained using depth information - frame by frame  \\
\includegraphics[width=0.16\textwidth]{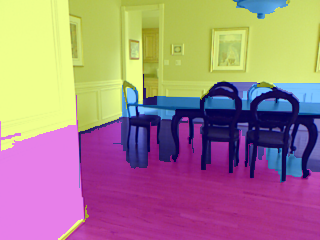}~
\includegraphics[width=0.16\textwidth]{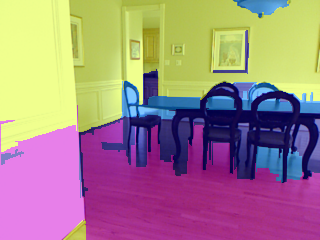}~
\includegraphics[width=0.16\textwidth]{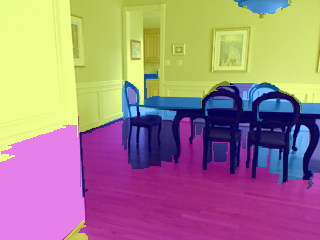}~
\includegraphics[width=0.16\textwidth]{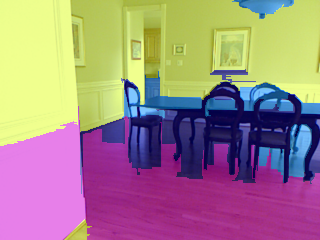}~
\includegraphics[width=0.16\textwidth]{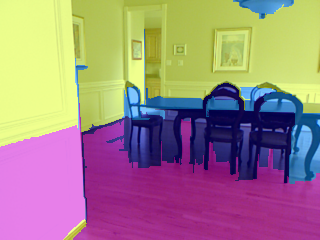}~
\includegraphics[width=0.16\textwidth]{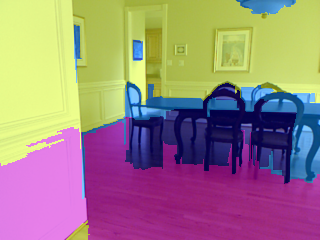}\\
(d) Results smoothed temporally using \cite{couprieCVPR2013}\\
\end{center}
\caption{Some results on video sequences of the NYU v2 depth dataset. Note that results (c,d) could be improved by using more training examples. Indeed, only a very small number in the labeled training examples exhibit a wall in the foreground.}
\label{fig:results_videos}
\end{figure*}

The predictions are computed frame by frame on the videos and are refined using temporally smoothed superpixels using \cite{couprieCVPR2013}. 
Two examples of results on sequences are shown at Figure \ref{fig:results_videos}.

A great advantage of our approach is its nearly real time capabilities. Processing a 320x240 frame takes 0.7 seconds on a laptop \cite{FarabetPAMI}. The temporal smoothing only requires an additional 0.1s per frame. 

 
\section{Conclusion}

Feature learning is a particularly satisfying strategy to adopt when
approaching a dataset that contains new image (or other kind of data)
modalities. Our model, while being faster and more efficient than
previous approaches, is easier to implement without the need to design
specific features adapted to depth information. Different clusterings
of object classes as the ones used in this work may be chosen,
reflecting this work's flexibility of applications. For example, using
the 4-classes clustering, the accurate results achieved with the
multi-scale convolutional network could be applied to perform
inference on support relations between objects. Improvements for
specific object recognition could further be achieved by filtering the
frequency of the training objects. We observe that the recognition of
object classes having similar depth appearance and location is
improved when using the depth information. On the contrary, it is better to use only
RGB information to recognize objects with classes containing high
variability of their depth maps. This observation could be used to combine the best results in function of the application. Finally, a number of techniques
(unsupervised feature learning, MRF smoothing of the convnet
predictions, extension of the training set) would probably help to
improve the present system.

\section{Acknowledgments}

We would like to thank Nathan Silberman for his useful input for handling the NYU depth v2 dataset, and fruitful discussions.

\bibliographystyle{plain}
\bibliography{iclr2013_depth}

\begin{thebibliography}{10}

\bibitem{berkeley}
B3do: Berkeley 3-d object dataset.
\newblock http://kinectdata.com/.

\bibitem{cornell}
Cornell-rgbd-dataset.
\newblock http://pr.cs.cornell.edu/sceneunderstanding/data/data.php.

\bibitem{Ciresan:2012f}
Dan~Claudiu Ciresan, Alessandro Giusti, Luca~Maria Gambardella, and J{\"u}rgen
  Schmidhuber.
\newblock Deep neural networks segment neuronal membranes in electron
  microscopy images.
\newblock In {\em NIPS}, pages 2852--2860, 2012.

\bibitem{Coup1208:Multi}
Camille Couprie.
\newblock Multi-label energy minimization for object class segmentation.
\newblock In {\em 20th European Signal Processing Conference 2012 (EUSIPCO
  2012)}, Bucharest, Romania, August 2012.

\bibitem{couprieCVPR2013}
Camille Couprie, Cl\'ement Farabet, and Yann LeCun.
\newblock Causal graph-based video segmentation, 2013.
\newblock arXiv:1301.1671.

\bibitem{cruz2012kinectTutoriel}
L.~Cruz, D.~Lucio, and L.~Velho.
\newblock Kinect and rgbd images: Challenges and applications.
\newblock {\em SIBGRAPI Tutorial}, 2012.

\bibitem{Dani04colorizationusing}
Anat~Levin Dani, Dani Lischinski, and Yair Weiss.
\newblock Colorization using optimization.
\newblock {\em ACM Transactions on Graphics}, 23:689--694, 2004.

\bibitem{FarabetICML2012}
Clement Farabet, Camille Couprie, Laurent Najman, and Yann LeCun.
\newblock {Scene Parsing with Multiscale Feature Learning, Purity Trees, and
  Optimal Covers}.
\newblock In {\em Proc.~of the 2012 International Conference on Machine
  Learning}, Edinburgh, Scotland, June 2012.

\bibitem{FarabetPAMI}
Clement Farabet, Camille Couprie, Laurent Najman, and Yann LeCun.
\newblock Learning hierarchical features for scene labeling.
\newblock {\em IEEE Transactions on Pattern Analysis and Machine Intelligence},
  2013.
\newblock In press.

\bibitem{Felzenszwalb04efficientgraph-based}
Pedro~F. Felzenszwalb and Daniel~P. Huttenlocher.
\newblock Efficient graph-based image segmentation.
\newblock {\em International Journal of Computer Vision}, 59:2004, 2004.

\bibitem{gould09_iccv}
Stephen Gould, Richard Fulton, and Daphne Koller.
\newblock {Decomposing a Scene into Geometric and Semantically Consistent
  Regions}.
\newblock In {\em IEEE International Conference on Computer Vision}, 2009.

\bibitem{imagenet}
Geoffrey~E. Hinton, Nitish Srivastava, Alex Krizhevsky, Ilya Sutskever, and
  Ruslan Salakhutdinov.
\newblock Improving neural networks by preventing co-adaptation of feature
  detectors.
\newblock {\em CoRR}, abs/1207.0580, 2012.

\bibitem{speech}
Navdeep Jaitly, Patrick Nguyen, Andrew Senior, and Vincent Vanhoucke.
\newblock Application of pretrained deep neural networks to large vocabulary
  speech recognition.
\newblock In {\em Proceedings of Interspeech 2012}, 2012.

\bibitem{Janoch11ICCVkinectworkshop}
Allison Janoch, Sergey Karayev, Yangqing Jia, Jonathan~T. Barron, Mario Fritz,
  Kate Saenko, and Trevor Darrell.
\newblock A category-level 3-d object dataset: Putting the kinect to work.
\newblock In {\em ICCV Workshops}, pages 1168--1174. IEEE, 2011.

\bibitem{Lecun1998gradient}
Y.~LeCun, L.~Bottou, Y.~Bengio, and P.~Haffner.
\newblock Gradient-based learning applied to document recognition.
\newblock {\em Proceedings of the IEEE}, 86(11):2278 --2324, nov 1998.

\bibitem{Lecun2004norb}
Yann LeCun, Fu-Jie Huang, and Leon Bottou.
\newblock {Learning Methods for generic object recognition with invariance to
  pose and lighting}.
\newblock In {\em Proceedings of CVPR'04}. IEEE, 2004.

\bibitem{liu2010}
Ce~Liu, Jenny Yuen, and Antonio Torralba.
\newblock {SIFT Flow: Dense Correspondence across Scenes and its Applications.}
\newblock {\em IEEE transactions on pattern analysis and machine intelligence},
  pages 1--17, August 2010.

\bibitem{torch}
C.~Farabet R.~Collobert, K.~Kavukcuoglu.
\newblock Torch7: A matlab-like environment for machines learning.
\newblock In {\em Big Learning Workshop (@ NIPS'11), Sierra Nevada, Spain},
  2011.

\bibitem{Ren2012NYUdepthCVPR}
Xiaofeng Ren, Liefeng Bo, and D.~Fox.
\newblock Rgb-(d) scene labeling: Features and algorithms.
\newblock In {\em Computer Vision and Pattern Recognition (CVPR), 2012 IEEE
  Conference on}, pages 2759 --2766, june 2012.

\bibitem{SocherEtAl2012:CRNN}
{Richard Socher and Brody Huval and Bharath Bhat and Christopher D. Manning and
  Andrew Y. Ng}.
\newblock {Convolutional-Recursive Deep Learning for 3D Object Classification}.
\newblock In {\em {Advances in Neural Information Processing Systems 25}}.
  2012.

\bibitem{Schulz2012learning}
Hannes Schulz and Sven Behnke.
\newblock Learning object-class segmentation with convolutional neural
  networks.
\newblock In {\em 11th European Symposium on Artificial Neural Networks
  (ESANN)}, 2012.

\bibitem{Silberman2011ICCVworkshop}
Nathan Silberman and Rob Fergus.
\newblock Indoor scene segmentation using a structured light sensor.
\newblock In {\em 3DRR Workshop, ICCV'11}, 2011.

\bibitem{Silberman:ECCV12}
Nathan Silberman, Derek Hoiem, Pushmeet Kohli, and Rob Fergus.
\newblock Indoor segmentation and support inference from rgbd images.
\newblock In {\em ECCV}, 2012.

\end{thebibliography}

\end{document}